\documentclass[sigconf]{acmart}
\usepackage[export]{adjustbox}
\usepackage{subfigure}
\usepackage{makecell}
\usepackage{titlesec}
\usepackage{amsmath,amsfonts}
\usepackage{algorithmic}
\usepackage{graphicx}
\usepackage{textcomp}
\usepackage{xcolor}
\usepackage{multirow}
\usepackage{hyperref}
\usepackage{booktabs}
\usepackage{hhline}
\usepackage[shortlabels]{enumitem}
\usepackage{soul}
\usepackage{fancyhdr}   
\usepackage{calc}
\usepackage{atbegshi}
\usepackage{tikz}

\usepackage{geometry}
\usepackage[utf8]{inputenc}

\usepackage{fancyhdr}
\newlength{\footeroffset}
\AtBeginDocument{%
  }

\definecolor{rubinered}{HTML}{CE0058}

\definecolor{green}{rgb}{0.0, 0.65, 0.31}

\definecolor{bleudefrance}{rgb}{0.19, 0.55, 0.91}

\definecolor{ao(english)}{rgb}{0.0, 0.5, 0.0}

\definecolor{violet}{HTML}{6a51a3}

\titlespacing{\section }{2pt}{3pt}{2pt}
\titlespacing{\subsection}{3pt}{2pt}{2pt}
\titlespacing{\subsubsection}{2pt}{2pt}{5pt}
\begin{document}

\title{Generating Virtual On-body Accelerometer Data from Virtual Textual Descriptions for Human Activity Recognition}

\author{Zikang Leng}
\email{zleng7@gatech.edu}
\affiliation{%
  \institution{Georgia Institute of Technology}
  \city{Atlanta, Georgia}
  \country{USA}}
\author{Hyeokhyen Kwon}
\email{hyeokhyen.kwon@dbmi.emory.edu}
\affiliation{%
  \institution{Emory University}
  \city{Atlanta, Georgia}
  \country{USA}}
\author{Thomas Plötz}
\email{thomas.ploetz@gatech.edu}
\affiliation{%
  \institution{Georgia Institute of Technology}
  \city{Atlanta, Georgia}
  \country{USA}}

\renewcommand{\shortauthors}{Leng et al.}

\begin{abstract}
\noindent
The development of robust, generalized models in human activity recognition (HAR) has been hindered by the scarcity of large-scale, labeled data sets. 
Recent work has shown that virtual IMU data extracted from videos using computer vision techniques can lead to substantial  performance improvements when training HAR models  combined with small portions of real IMU data.
Inspired by recent advances in motion synthesis from textual descriptions and connecting Large Language Models (LLMs) to various AI models, we introduce an automated pipeline that first uses ChatGPT to generate diverse textual descriptions of activities. 
 These textual descriptions are then used to generate 3D human motion sequences via a motion synthesis model, T2M-GPT,  
 and later converted to streams of virtual IMU data. 
We benchmarked our approach on three HAR datasets (RealWorld, PAMAP2, and USC-HAD) and demonstrate that the use of virtual IMU training data generated using our new approach leads to significantly improved HAR model performance compared to only using real IMU data. 
Our approach contributes to the growing field of cross-modality transfer methods and illustrate how HAR models can be improved through the generation of virtual training data that  do not require any manual effort.

\end{abstract}



\maketitle

\thispagestyle{fancy}
\fancyfoot{} 
\pagenumbering{arabic}
\pagestyle{fancy}
\fancyhf{}
\renewcommand{\headrulewidth}{0pt}
\AtBeginShipout{\AtBeginShipoutAddToBox{%
  \begin{tikzpicture}[remember picture, overlay, red]
    \node[anchor=south, font=\LARGE] at ([yshift=15mm]current page.south) {This manuscript is under review. Please write to zleng7@gatech.edu for up-to-date information};
  \end{tikzpicture}%
}}

\section{Introduction}
\noindent
The development of accurate and robust predictive models for human activity recognition (HAR) is essential for, e.g., monitoring fitness, analyzing health-related behavior, and improving industrial processes \cite{bachlin2010wearable, chavarriaga2013opportunity, Liaqat2019, Stiefmeier2008}. 
However, one of the major challenges in HAR research is the scarcity of labeled activity data, which hinders the effectiveness of supervised learning methods \cite{chen2021sensecollect}. 


To address this challenge, researchers have explored innovative methods for acquiring labeled data that are more flexible and cost-effective. One such method is the use of virtual IMU data generation. In recent years, effective \textit{cross-modality transfer} approaches \cite{kwon2020imutube, kwon2021approaching, kwon2021complex} have been utilized to extract virtual IMU data from 2D RGB videos of human activities. Virtual IMU data can expand training datasets for motion exercise recognition and can be used to build personalized HAR systems that meet the diverse needs of individual users \cite{Xia2022}. By leveraging the advantages of virtual IMU data, researchers can improve the accuracy and robustness of HAR models and facilitate the widespread adoption of sensor-based HAR in a variety of domains.

In this work, we share a similar motivation and present a method that can \textit{generate diverse textual descriptions of activities that can then be converted to streams of virtual IMU data}. 
In our automated pipeline, the name of an activity is first passed to ChatGPT to automatically generate textual prompts that describes a person doing the activity, for example: 

\begin{quote}
    \textbf{Activity (user specified):} Running

    \textbf{ChatGPT prompt 1  (generated):} A sprinter races towards the finish line, narrowly beating their competition.

    \textbf{Chat GPT prompt 2 (generated):} A person runs towards their love interest in a romantic reunion.

    \textbf{\ldots}
\end{quote}
The generated textual prompts are then used to generate 3D human motion using a motion synthesis model, which can then be converted to streams of virtual IMU data. By using ChatGPT to generate the diverse textual descriptions of activities, we can generate virtual IMU data that capture the different variations of how activities can be performed. With ChatGPT, no prompt engineering is needed and essentially unlimited amounts of virtual IMU data can be generated. 

\begin{figure*}[t]
    \centering
    \includegraphics[width=.9\linewidth]{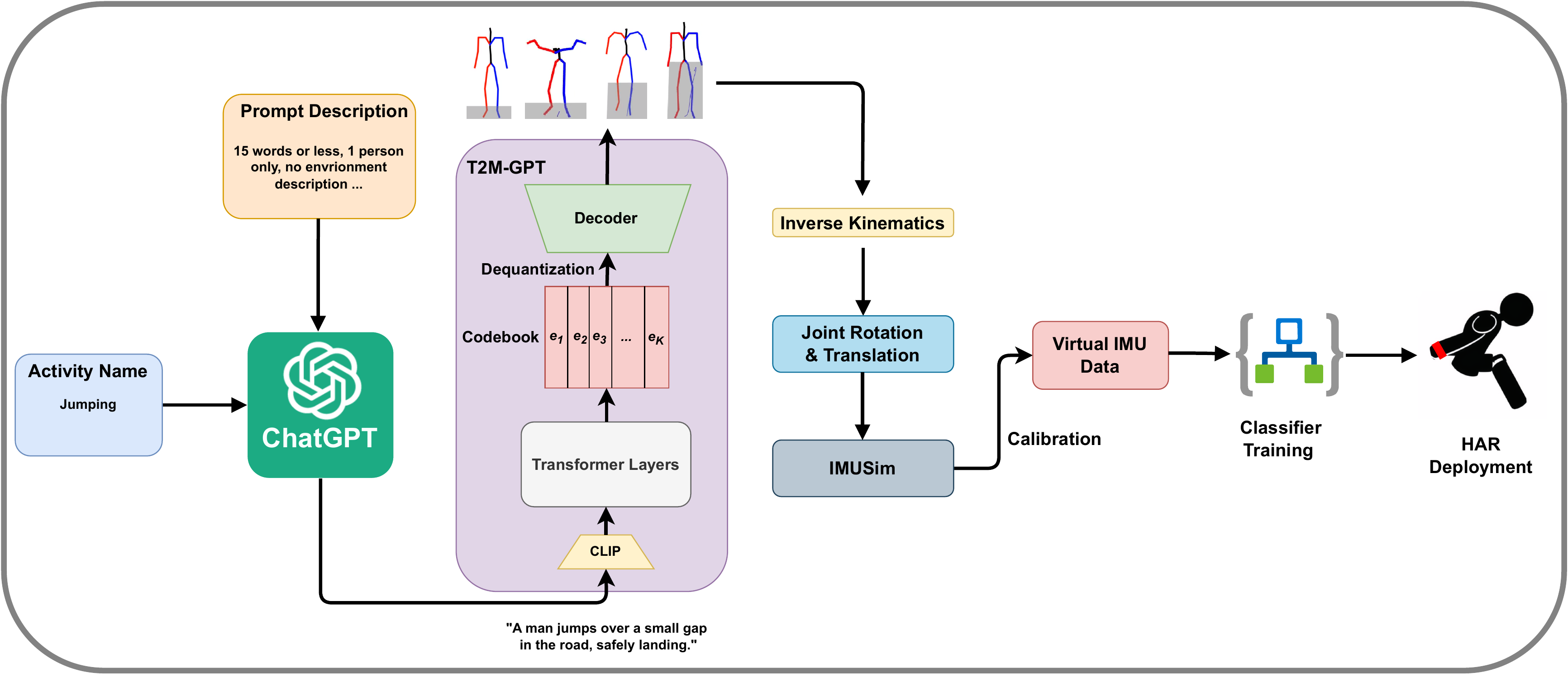}
    \caption{Overview of the proposed approach. 
    The name of the desired activity, general description of prompts, and example prompts are provided to ChatGPT for prompts generation. 
    Using the generated prompts, T2M-GPT generates 3D human motion sequence. 
    Using the motion sequence, the joint rotations and translations are estimated through inverse kinematics \cite{Yamane2003ik}. 
    Using the estimated joint rotations and translations, IMUSim calculates the virtual IMU data at each joint. 
    After calibrating the virtual IMU data  with a small amount of real IMU data, the virtual IMU data can be used to train a depolyable classifier.
    }
    \label{fig:t2imu}

    \vspace{-0.2in}
\end{figure*}

The contributions of this paper are two-fold:
\begin{enumerate}
\item 
We leverage ChatGPT's natural language generation capabilities to automaticelly generate textual descriptions of activities, which are then used in conjunction with motion synthesis and signal processing techniques to generate virtual IMU data streams. By using this approach, we can significantly reduce the time and cost required for data collection, while covering a wide range of activity variations.

\item We evaluate our approach on three standard HAR datasets -- Realworld, Pamap2, and USC-HAD -- and demonstrate the overall effectiveness through improved activity recognition results across the board for models that utilize virtual IMU data generated through our approach.
\end{enumerate} 

The results of our approach are significant -- they contribute to the growing field of cross-modality transfer that promises to alleviate the much lamented lack of annotated training data in HAR -- thereby virtually requiring no manual effort at all. 

\section{Related Work}
\noindent
\textit{Virtual IMU Data Generation: } Recently, IMUTube \cite{kwon2020imutube} was introduced to extract virtual IMU from 2D RGB videos. IMUTube uses computer vision methods such as 2D/3D pose tracking to extract the 3D human motion in the given video. The extracted 3D human motion is used to estimate 3D joint rotations and global motion, which is then used calculate the virtual IMU data. Previous studies \cite{kwon2021approaching, kwon2021complex} have shown that the extracted virtual IMU data led to improved model performance when mixed with the real IMU data and allowed for effective training of more complex models. 

To improve the quality of the extracted virtual IMU data, Xia \textit{et al.} \cite{Xia2022} proposed a spring-joint model to augment the extracted virtual acceleration signal and trained a classifier on the augmented virtual IMU data to recognize reverse lunge, warm up, and high knee tap. Vision-based systems such as IMUTube is limited by the quality of the video. In order for the extracted virtual IMU data to be of suitable quality, the input video should exhibit little to no camera egomotion and only include people performing the desired activity. Hence, selecting videos of good quality can be time-consuming. Since our system is text-based, the time-consuming process of selecting videos is eliminated. 

\textit{Text-driven Human Motion Synthesis:} The goal of text-driven Human Motion Synthesis is to generate 3D human motion using textual descriptions. With the recently released HumanML3D \cite{Guo_2022_CVPR}, the current largest 3D human motion dataset with textual descriptions, numerous models have been introduced that can produce significantly more realistic human motion sequences than previous models. MDM \cite{tevet2022human}, MLD \cite{chen2023mld}, and MotionDiffuse \cite{zhang2022motiondiffuse} are three recently introduced diffusion-based models. In this work, we use T2M-GPT \cite{zhang2023generating} as the motion synthesis model for our system. T2M-GPT is based on Vector Quantized Variational Autoencoders (VQ-VAE) \cite{oord2017vq-vae} and Generative Pre-trained Transformer (GPT) \cite{Radford2018ImprovingLU, Vaswani2017attn}. The model can be viewed as two parts. The first part is an encoder that learns the mapping between human motion sequence and discrete code indices, which corresponds to latent vectors in a codebook. The second part is a transformer that learns to generate code indices from embedded textual prompts. At inference, the generated code indices are mapped to human motion using a learned decoder. 

\textit{Large Language Models: } Large Language Models (LLMs) such as PaLM \cite{chowdhery2022palm}, LLaMA \cite{touvron2023llama}, GPT-3 \cite{brown2020gpt3}, and ChatGPT (built upon InstructGPT \cite{Ouyang2022instruct}) have attracted enormous attentions for their superior performances in many natural language processing (NLP) tasks. However, LLMs alone cannot solve complex AI tasks that require processing information from multiple modalities such as vision. Recently, Visual ChatGPT \cite{wu2023visual} and HuggingGPT \cite{shen2023hugginggpt} were introduced to tackle complex multi-modal tasks. Both use ChatGPT as a controller that can divide user input into sub-tasks and select the relevent AI model from a pool of models to solve the complex task. Inspired by this idea, we use ChatGPT as a prompt generator to generate diverse textual descriptions for activities that are then used as input for the motion synthesis model in our system. 

\section{Generating Virtual IMU Data from Virtual Textual Descriptions}
\noindent
\noindent
The key idea of our approach lies in generating a wide range of diverse textual descriptions for a given activity, and to then feed those textual descriptions into a motion synthesis model that is connected to a virtual IMU data generation pipeline. 
Fig.\ \ref{fig:t2imu} provides an overview of the developed approach. 
Human activities are inherently variable; a person can walk happily, confidently, quickly, or in many other ways. This variability is reflected in the IMU data collected by wearable sensors, which must be accurately represented in the training data to ensure HAR models generalize well. 
We address this challenge by employing ChatGPT to--automatically--create detailed and varied textual descriptions of activities, which then serve as prompts for 3D human motion synthesis. 

During prompt generation, the activity name, few example textual descriptions (not activity specific), general description of the desired prompts are provided to ChatGPT. The example textual descriptions serves as few-shot examples that ChatGPT can learn from. The prompt description is provided to help align ChatGPT's output with our desired prompts. Some descriptions that we used include: \textit{prompts should be 15 words or less; prompts should only include a single person performing the activity; prompts should not contain extensive description of the environment. }\footnote{Scripts, generated prompts, and virtual IMU data will be shared upon publication.} 

The generated prompts are then fed into the motion synthesis model, T2M-GPT \cite{zhang2023generating} trained on HumanML3D, to generate 3D human motion sequences. 
To do so, CLIP \cite{Radford2021clip}, a pre-trained text encoder, first extracts the text embedding from the prompt. 
Using this, a learned transformer generates code indices autoregressively until an end token is generated. 
The sequence of code indices is de-quantized into latent vectors by looking up the corresponding vector in the codebook for each index. 
Lastly, a learned decoder maps the sequence of latent vectors to 3D human motion sequence, represented as a sequence of 22 joints' positions. 

\begin{table}
\centering
    \caption{Real and virtual IMU datasets size for the three HAR datasets we used for evaluation. 
    }
    \vspace*{-1.0em}
    \begin{adjustbox}{width=0.55\columnwidth,center}
    \begin{tabular}{c|c|c}
        Dataset & Real Size & Virtual Size \\
         \hline\hline
        RealWorld & 1,107 min & 41 min\\ 
         \hline
         PAMAP2  &  322 min & 68 min \\
         \hline
          USC-HAD &  469 min & 69 min \\
    \end{tabular}
    \end{adjustbox}
    \label{tab:dataset}
    \vspace*{-0.5em}
\end{table}

We estimate each joint's rotation with respect to the parent joint and the root joint's (pelvis) 
translation using inverse kinematics \cite{Yamane2003ik} with the joints' positions and the skeleton's hierarchical structure as input.
IMUSim \cite{Yound2011imusim} is then used to calculate the joint's acceleration movement and angular velocity using the estimated local joints' rotations and root translation. 
This allows us to extract virtual IMU data from 22 on-body sensor locations. Additionally, IMUSim introduces noises to the generated virtual IMU data to simulate the noises that real IMU data typically exhibit.

Inevitably there will be a domain gap between the virtual IMU data's domain (source) and the real IMU data's domain (target) due to potential differences in coordinate systems, sensor orientations and placements, and the size of real human and virtual skeleton. 
We employ domain adaptation to bridge the gap between the two domains. 
Following \cite{kwon2020imutube}, we perform a distribution mapping between the virtual IMU data and the real IMU data  using the rank transformation approach \cite{Conover1981rank}. 
To calibrate the virtual IMU data, only a small amount of real IMU data is needed. 

After calibration, the process of virtual IMU data generation is complete. The extracted virtual IMU data can then be used to train a HAR model either alone or in combination with some real IMU data. Lastly, the trained model is  deployed in the real world.

\section{Experimental Evaluation}

\noindent
We evaluated the effectiveness of our approach in a set of experiments where we train activity recognizers for benchmark recognition tasks and analyze the performance (F1 scores) for scenarios where only real, only virtual, and mixtures of real and virtual training data are used (similar to previous work, e.g., \cite{kwon2020imutube,kwon2021approaching,kwon2021complex}).

\subsection{Datasets}
\noindent
\textit{Real IMU Dataset:\quad} 
To evaluate the value of the virtual IMU data generated by our proposed approach, we use the RealWorld \cite{Sztyler2016realworld}, PAMAP2 \cite{Reisspamap}, and USC-HAD \cite{Zhang2012usc} datasets (details in  \autoref{tab:dataset}). 

 \begin{table}[t]
    \centering
    \footnotesize
    \caption{Model performances (Macro F1) for the experimental evaluation of our approach for  the three HAR datasets. The best performance within each scenario is highlighted in bold. }
    \vspace*{-1em}
    \begin{adjustbox}{width=1\columnwidth,center}
    \begin{tabular}{c|c|c|c}
         Dataset & PAMAP2 & RealWorld & USC-HAD\\
         \hline\hline
         Real & $0.659 \pm 0.003$ & $0.715 \pm 0.011$ & $0.478 \pm 0.002$\\
         \hline
         Virtual & $0.628 \pm 0.003$ & $0.746 \pm 0.003$   & $0.448 \pm 0.003$ \\
         \hline
         Real+Virtual & $\textbf{0.699} \pm 0.004$ & $\textbf{0.770} \pm 0.004$ & $\textbf{0.486} \pm 0.003$ \\
        \hline
    \end{tabular}
    \end{adjustbox}
    \label{tab:results}
    \vspace*{-1em}
\end{table}

RealWorld contains IMU data collected from 15 subjects performing eight locomotion-style activities in a naturalistic setting, presenting reasonable variability in how activities are performed. 
The eight activities are: \textit{climbing up stairs, climbing down stairs, jumping, lying, running, sitting, standing, and walking.} 
While performing the activities, the subject wore sensors at seven body locations: \textit{forearm, head, shin, thigh, upper arm, waist, and chest}.

For  PAMAP2 \cite{Reisspamap}, we use all twelve activities from the original protocol: \textit{lying, sitting, standing, walking, running, cycling, Nordic walking, ironing, vacuum cleaning, rope jumping, ascending stairs, and descending stairs}.
The activities were performed by nine subjects. 
Subject nine's data only contained rope jumping, so we did not use subject nine in the experiment. 
The subjects wore the sensors at three body locations: \textit{forearm, chest, and ankle.} 

USC-HAD contains IMU data of 14 subjects performing twelve activities: \textit{walking forward, walking counter-clockwise, walking clockwise, climbing upstairs, climbing downstairs, running, jumping, sitting, standing, sleeping, riding an ascending elevator, riding a descending elevator.} 
The sensor was attached to the subject's \textit{right hip.} 
All real IMU datasets were downsampled to 20 Hz to match the virtual IMU datasets.

\textit{Virtual IMU Dataset:\quad} 
To generate the virtual IMU dataset, we used our system to generate 50 clips of virtual IMU data for each activity. 
Each clip corresponds to a different--automatically generated--prompt from ChatGPT,
The length of the clips ranges from five to ten seconds, and the exact length of the clip depends on when the transformer generates the end token, which in turn depends on the textual prompt. 
The virtual IMU data was extracted from joint locations of the virtual skeleton that were selected to be physically closest to the sensor locations on the subjects.



\begin{figure*}[t]
    \centering
    \includegraphics[width=.9\linewidth]{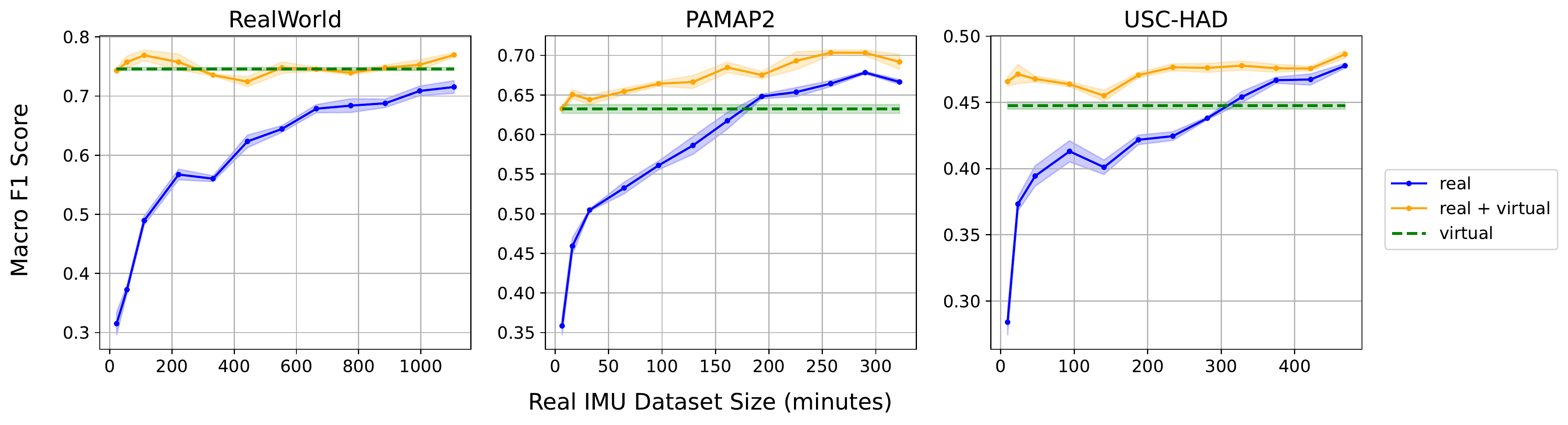}
    \vspace{-0.1in}
    \caption{ Model performance on RealWorld \cite{Sztyler2016realworld}, PAMAP2 \cite{Reisspamap}, and USC-HAD \cite{Zhang2012usc} datasets when different amount of real IMU data are used for training. The amount of virtual IMU data used remains the same. 
    }
    \label{fig:vary_imu}

    \vspace{-0.2in}
\end{figure*}

\subsection{Classifier Training}
\noindent
We use a standard Random Forest classifier as our backend. 
Sliding windows of 2 seconds duration and with 50\% overlap are used to segment the real and virtual IMU data. 
ECDF features \cite{hammerla2013preserving} (15 components) are extracted from the windows for training. 
We train a classifier only on the real IMU data to establish a baseline. Additionally, we trained a classifier on only virtual IMU data and another classifier on both real and virtual IMU data. 
Only the accelerometry signal is used since \cite{kwon2020imutube} showed that the inclusion of angular velocity is not beneficial. 
For evaluation, we performed leave-one-subject-out cross-validation on the real IMU dataset with a test set of 1 subject in each fold. 
The training set is not used when training a classifier only on virtual IMU data. 
We report macro F1 scores averaged across all folds over three runs. 

To evaluate the benefit of the virtual IMU data when different amounts of real IMU data is available, we varied the amount of real IMU data used for training. Starting with $2\%$ of the available to real IMU data for training, we gradually increased the size of the real IMU dataset for training. 
The virtual IMU dataset and the testing dataset is left unchanged. 

\subsection{Results}


\noindent
Results are listed in \autoref{tab:results}.
 The classifier trained on both real and virtual IMU data shows $6.1 \%, 7.7 \%, $  and $ 1.7 \%$ relative improvement in F1 score compared to a classifier trained only on real IMU data for the PAMAP2, RealWorld, and USC-HAD datasets respectively.  Furthermore, on the RealWorld dataset, we observe that the classifier trained on only virtual IMU data outperforms the classifier trained on real IMU data. We find this surprising because the size of the virtual IMU dataset is less than 4\% of the size of the real IMU dataset. We attribute this performance improvement to the diverse textual prompts that ChatGPT generated, which led to a diverse set of virtual IMU clips. Using such a diverse training data, the model learns to recognize the many variations of each activity. 


\autoref{fig:vary_imu} shows the model performances when varying amount real IMU data is used for training. We observe that the classifier trained on both real and virtual IMU data consistently outperform the classifier trained only on real IMU data for varying amount of real IMU data. The performance improvement is especially apparent when the size of the real IMU dataset is greatly reduced. This shows the use of virtual IMU data for training is exceptionally beneficial when the amount of available real IMU data is limited.




\section{Discussion}

\begin{figure}
    \centering
    \vspace{-0.09in}
    \subfigure[]{\includegraphics[width=.6\linewidth]{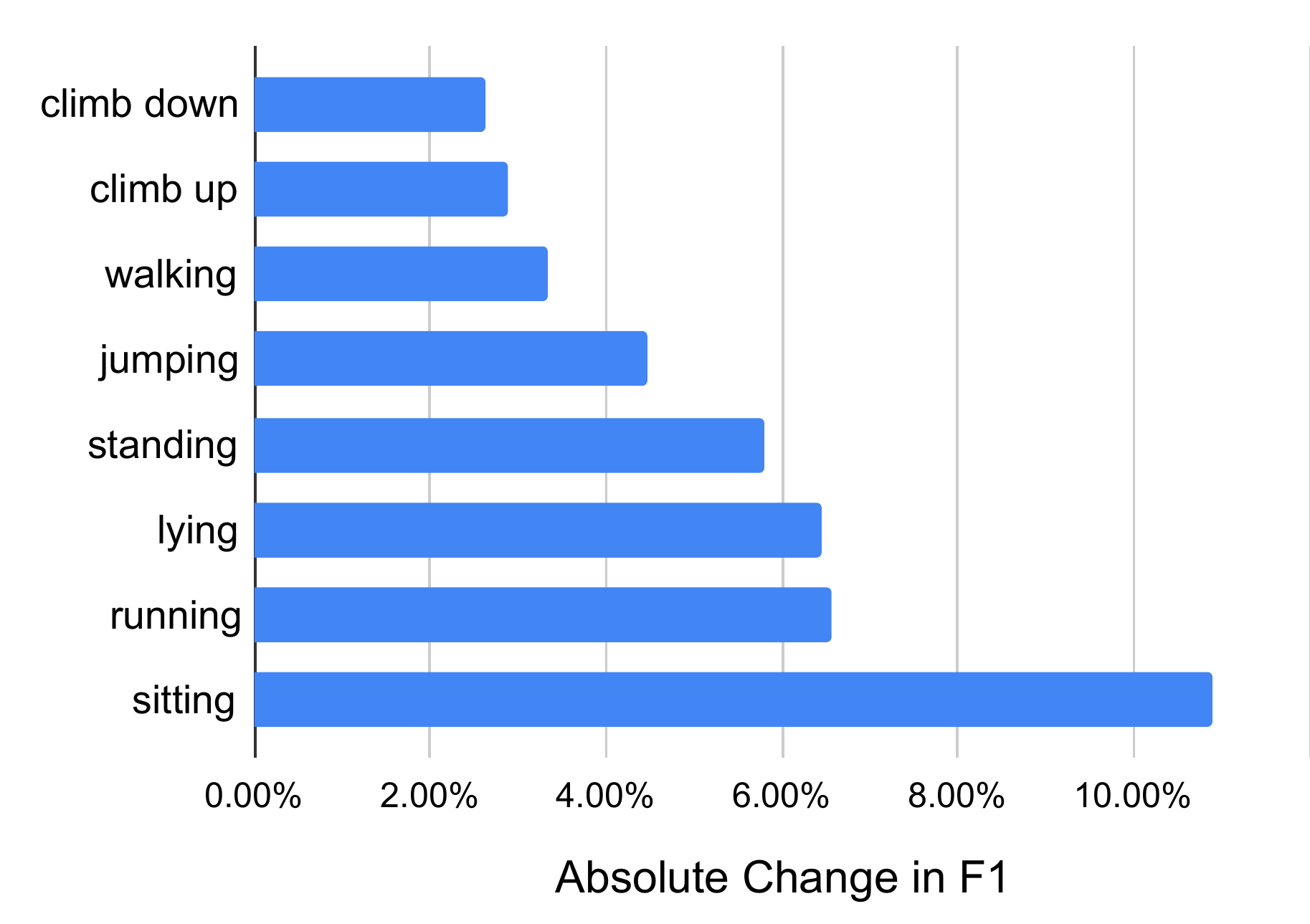}
    \label{fig:f1_per_cls}
    }
    \subfigure[]{\includegraphics[width=.33\linewidth]{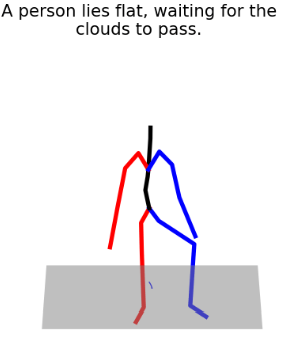}
    \label{fig:lying_vis}
    }
    \vspace*{-1.7em}
        \caption{(a) Differences in F1 score for each activity between the classifier trained on only real IMU data and the classifier trained on both real and virtual IMU data evaluated on the RealWorld \cite{Sztyler2016realworld} dataset. (b) Example where the motion synthesis model confused waiting with lying. }
    \label{fig:discussion}

    
\end{figure}


\noindent
The experimental evaluation demonstrates the effectiveness of our proposed approach.
In this section we explore current limitations and outline directions for future research that could further enhance the utility of our method.

First, the pipeline will only be able to generate virtual IMU data for activities that are described in the HumanML3D dataset. 
If the prompt contains activities that are not captured by the HumanML3D dataset, our pipeline will fail to generate realistic virtual IMU data for the activity.
One potential solution would be to extend the HumanML3D dataset with new activities. 
A cost-effective method for extension would be to use computer vision techniques such as 3D human pose estimation \cite{ye2023slahmr} on existing videos to extract the human motion sequence for the new activities. 

Second, the motion synthesis model sometimes confuses closely related activities or two verbs in the same prompt. 
For instance, T2M-GPT sometimes generates a motion sequence for climbing up the stairs when the input prompt is for climbing down the stairs and vice versa. 
As per Fig.~\autoref{fig:f1_per_cls} climbing up and down stairs gained the least increase in per class f1 score from the addition of virtual IMU data. 
Additionally, T2M-GPT sometimes confuses another verb in the prompt for the activity. 
As shown in Fig.~\autoref{fig:lying_vis}, T2M-GPT confuses "waiting" with "lies", which causes the generated  motion sequence to be more similar to sitting than lying. 
A potential solution for this problem is prompts weighting (often used in text-to-image generation \cite{rombach2021highresolution}), giving more weights to the activity-related parts of the prompt, which allows the motion synthesis model to focus more on the activity.

We plan to explore ways to further increase the diversity of the generated virtual IMU data. 
We will test diffusion-based motion synthesis models \cite{tevet2022human, chen2023mld, zhang2022motiondiffuse}, which generate more diverse motion sequences for similar prompts and use motion style transfer \cite{aberman2020unpaired} to apply different motion styles to the generated motion sequences.

\section{Conclusion}
\noindent
We have introduced a method that uses ChatGPT to generate virtual textual descriptions, which are subsequently used to generate 3D human motion sequence and later streams of virtual IMU data. 
We have demonstrated the effectiveness of our approach to generate virtual IMU data through HAR experiments on three benchmark datasets: RealWorld, PAMAP2, and USC-HAD. 
Virtual IMU data generated through our approach can be used for significantly improving the recognition performance of HAR models -- bringing $1.7 \% - 7.7 \%$ relative improvement in performance on the three benchmark datasets -- thereby not requiring any additional manual effort. 

\bibliographystyle{ACM-Reference-Format}
\bibliography{./bibs/virtual_imu,./bibs/motion_synthesis, ./bibs/har, ./bibs/llm}

\end{document}